\documentclass{article}

\usepackage{graphicx}

\usepackage{hyperref}

\usepackage[accepted]{icml2020}

\usepackage{bm}
\usepackage{mathtools}
\usepackage{amsmath}
{
      \newtheorem{assumption}{Assumption}
      
}
\usepackage{amsfonts}
\usepackage{amssymb}

\usepackage{booktabs}
\usepackage{tikz}

\usepackage{subcaption}

\usepackage{xcolor}

\definecolor{myblue}{RGB}{195,228,247}
\definecolor{mygreen}{RGB}{198,232,175}
\definecolor{mypurple}{RGB}{208,186,207}
\definecolor{mypink}{RGB}{253,188,175}

\DeclareMathOperator*{\argmin}{arg\,min}

\newcommand\independent{\protect\mathpalette{\protect\independenT}{\perp}}
\def\independenT#1#2{\mathrel{\rlap{$#1#2$}\mkern2mu{#1#2}}}

\icmltitlerunning{Causal Coordinated Concurrent Reinforcement Learning}

\begin{document}

\twocolumn[
\icmltitle{Causal Coordinated Concurrent Reinforcement Learning}



\icmlsetsymbol{equal}{*}

\begin{icmlauthorlist}
\icmlauthor{Tim Tse}{h}
\icmlauthor{Isaac Chan}{h}
\icmlauthor{Zhitang Chen}{h}
\end{icmlauthorlist}

\icmlaffiliation{h}{Huawei Noah's Ark Lab}

\icmlcorrespondingauthor{Tim Tse}{tim.tse@huawei.com}
\icmlcorrespondingauthor{Isaac Chan}{isaac.chan@huawei.com}
\icmlcorrespondingauthor{Zhitang Chen}{chenzhitang2@huawei.com}

\icmlkeywords{Machine Learning, ICML}

\vskip 0.3in
]



\printAffiliationsAndNotice{}  

\begin{abstract}
In this work, we propose a novel algorithmic framework for data sharing and coordinated exploration for the purpose of learning more data-efficient and better performing policies under a concurrent reinforcement learning (CRL) setting. In contrast to other work which make the assumption that all agents act under identical environments, we relax this restriction and instead consider the formulation where each agent acts within an environment which shares a global structure but also exhibits individual variations. Our algorithm leverages a causal inference algorithm in the form of Additive Noise Model - Mixture Model (ANM-MM) in extracting model parameters governing individual differentials via independence enforcement. We propose a new data sharing scheme based on a similarity measure of the extracted model parameters and demonstrate superior learning speeds on a set of autoregressive, pendulum and cart-pole swing-up tasks and finally, we show the effectiveness of diverse action selection between common agents under a sparse reward setting. To the best of our knowledge, this is the first work in considering non-identical environments in CRL and one of the few works which seek to integrate causal inference with reinforcement learning (RL).\end{abstract}

\section{Introduction}
We address the problem of concurrent reinforcement learning (CRL) wherein each agent acts within an environment sharing a global structure but also exhibits individual variabilities specific to an agent. Concretely, consider the setting where multiple reinforcement learning (RL) agents are performing the classical pendulum swing-up task but with the slight modification in which each environment has a constant strength wind. The wind exerts a horizontal force on the pendulum which may or may not be the same across the environments. Intuitively, one can imagine that it is possible, for example, to share data amongst the agents and to coordinate their actions for the purpose of accelerating the learning of each controller. However, due to the presence of non-identical environments, it is not immediately clear how one may go about this.

This work seeks to address these research questions; specifically, we aim to investigate an algorithm to 1) extract and separate the variations across a set of MDPs, 2) leverage the knowledge of the similarities of environments to provide a better informed approach to policy and/or state-action value function learning and 3) coordinate the selection of actions for a set of agents so as to accelerate the exploration of their state space. To this end, we leverage ANM-MMs \cite{Hu:2018:CIM:3327345.3327427}  which is a recently proposed algorithm in causal inference that performs latent model parameter extraction via independence enforcement. Mechanism clustering is shown to be equivalent to clustering over the extracted latent parameters and in this work, we apply expectation--maximization (EM) \cite{Dempster77maximumlikelihood} to determine a soft similarity metric that is used to inform a data-sharing scheme for CRL. Finally, we introduce a simple yet effective sampling-based coordinated action selection heuristic which we show to be very effective under the sparse reward setting when coupled with our proposed algorithm.

In contrast to other causal inference methods based on functional models such as ANM \cite{NIPS2008_3548}, LiNGAM \cite{Shimizu:2006:LNA:1248547.1248619}, PNL \cite{Zhang:2009:IPC:1795114.1795190} and IGCI \cite{Janzing:2010:CIU:1870245.1870271} which assume a single causal model for all observations, we opt for the ANM-MM in this work due to its generalizing assumption that the observed data is generated from multiple sources with varying causal models in alignment with our goal of modelling concurrent MDPs which may be perturbed by multiple sources. There has been surprisingly little work in the area of CRL despite their applicability to important domains such as web services \cite{pmlr-v28-silver13} and robots concurrently learning how to perform a task \cite{DBLP:journals/corr/GuHLL16}. The work of \cite{pmlr-v28-silver13} and \cite{DBLP:journals/corr/GuHLL16} only considers data-gathering in parallel and does not address coordinated action selection. While the work of \cite{DBLP:journals/corr/abs-1802-01282} establishes essential conditions for efficient coordinated exploration and their follow-up work \cite{DBLP:journals/corr/abs-1805-08948} demonstrates the scalability of their algorithm to a continuous control problem, their frameworks assume identical MDPs across all concurrent instances. In this paper, we address data-gathering, coordinated action selection, and non-identical MDPs for CRL.

The remainder of the paper is organized as follows: first we provide some preliminaries, then we describe our proposed model. 
Next, we present experiments on autoregressive, sparse autoregressive, pendulum and cart-pole swing-up tasks and finally, we will conclude.
\section{Background and Related Work}
In this section, we provide some background to RL, concurrent RL, and ANM-MM and also discuss the related work seed sampling.
\subsection{Reinforcement Learning}
Reinforcement learning (RL) enables an agent with a goal to learn by continuously interacting with an environment. At each time step, the agent exists in a given \textit{state} within the environment and the agent makes an \textit{action} on the environment where the environment then provides feedback to the agent with the \textit{next state} and \textit{reward}. The above interaction can be modeled by a Markov decision process (MDP) consisting of $s_t \in S$ the set of states, $a_t \in A$ the set of actions, $P(r_t| s_t, a_t)$ the reward function which maps the reward that the agent receives in state $s_t$ taking action $a_t$, $P(s_{t+1}|s_t,a_t)$ the transition function which defines the probability of transitioning to state $s_{t+1}$ while taking action $a_t$ in state $s_t$ and $\gamma \in [0,1]$ the discount factor which sets the priority of short versus long term rewards. The goal of the agent is to find a policy $\pi: S \rightarrow A$ such that the expected discounted cumulative rewards $\mathbb{E}[\sum_{t=0}^T \gamma^t r_t | a_t=\pi(s_t)]$ in the lifetime $T$ of the agent is maximized.
\subsection{Concurrent Reinforcement Learning}
One extension to the RL framework is the concurrent RL formulation where instead of having one agent, there exists $N$ RL agents that are interacting in their own environments. There are benefits to be exploited, as agents may potentially share data and/or coordinate their exploration to learn more data-efficient and/or better performing policies. Much of the work in concurrent RL considers the scenario where each agent acts in identical environments (i.e., identical MDPs). In this work, we relax this restriction and instead assume that the state transition model and the reward function of different agents are not identical but share similarities. More concretely, we make the assumption that the state transition models of all agents have the same functional form which is parameterized by $\bm{\xi}$ and their reward functions share the same functional form parameterized by $\bm{\omega}$, i.e., \begin{align} P_n(s_{t+1} | s_t, a_t) &= P(s_{t+1} | s_t, a_t; \bm{\xi}_n), \\ R_n(s_t, a_t) &= R(s_t, a_t; \bm{\omega}_n) \end{align} for each $n^\text{th}$ agent. We stress that $\bm{\xi}$ and $\bm{\omega}$ are unknown, rendering the task of CRL with heterogeneous environments difficult.
\subsection{Additive Noise Model - Mixture Model}
\textbf{ANM-MM} Additive Noise Model - Mixture Models \cite{Hu:2018:CIM:3327345.3327427} is defined as a mixture of causal models of the same causal direction between two random variables $X$ and $Y$ where all the models share the same form given by the following Additive Noise Model (ANM) \cite{NIPS2008_3548}: \begin{equation}Y = f(X;\theta) + \epsilon \end{equation} where $X$ denotes the cause, $Y$ denotes the effect, $f$ is a nonlinear function parameterized by $\theta$ and $\epsilon \independent X$ is the noise. The goal of the ANM-MM model is to extract the model parameter $\theta$ and this is done by mapping $X$ and $Y$ to a latent space using Gaussian process partially observable model (GPPOM) and then incorporating the Hilbert-Schmidt independence criterion (HSIC) \cite{Gretton:2005:MSD:2101372.2101382} on the latent space of GPPOM to estimate the model parameter $\theta$. By assuming a Gaussian prior over the model parameters, the method proposes minimizing a log-likelihood loss function subject to a HSIC regularization term for extracting the latent parameters of the model. To identify the underlying data generating mechanisms, the authors propose a mechanism clustering technique amounting to conducting \textit{k}-means clustering directly on the extracted latent parameters.
\subsection{Seed Sampling}
In this section, we review seed sampling \cite{DBLP:journals/corr/abs-1805-08948} which is a recently proposed algorithm for coordinated exploration for CRL. We stress, however, that the problem formulation of seed sampling is different from ours in that the former assumes each concurrent MDP is identical. Notwithstanding, our problem formulation is niche to the extent where, to the best of our knowledge, there are no current works which exactly address our formulation and thus we settle with seed sampling which, in our opinion, is the closest alternative baseline.

Seed sampling is an extension of posterior sampling (aka Thompson Sampling \cite{10.1093/biomet/25.3-4.285}) for reinforcement learning (PSRL) \cite{Strens:2000:BFR:645529.658114} to the parallel agent setting satisfying properties of efficient coordinated exploration \cite{DBLP:journals/corr/abs-1802-01282}. The main idea is that  each agent $k$ possesses a sampled seed $\omega_k$ which is intrinsic to the agent and differentiates how agent $k$ perceives the shared data buffer $\mathcal{B}$. The seed $\omega_k$ remains fixed throughout the course of learning. One form of seed sampling involves each agent independently and randomly perturbing the observations in $\mathcal{B}$ via different noise term $z_k$ determined by seed $\omega_k$ which induces diversity by creating modified training sets from the same history among the agents. The presence of independent random seeds encourage diverse exploration while a consistent seed ensures a certain degree of commitment from each agent. We refer the reader to the original paper for further details on seed sampling.

\section{Algorithmic Framework}
Our algorithm can naturally be divided into two parts where in the first part, we apply the independence enforcement of ANM-MMs to extract the model parameters. Next, we perform soft clustering on said model parameters in the form of clustering with Gaussian Mixture Model (GMM) where then the similarity features given by the clustering model is used to dictate the sharing of data and the coordination of actions. We detail each constituent below. Algorithm \ref{ouralgo} provides a pseudocode summary of the overall algorithm.

Given $N$ RL agents, our work seeks to determine the model parameters $\theta$ of each agent given a cause $X$ and a resulting effect $Y$. We make note that in a significant bulk of literature in causal inference, the proposed algorithms seek to determine a causal direction $X \longrightarrow Y$ using sampled observations $\bm{x}_i$ and $\bm{y}_i$. Perhaps an intuitive extension of this model to RL data would involve finding model parameters of each agent using individual transition tuples $(s_t, a_t, r_t, s_{t+1})_n$ for each $n^\text{th}$ agent. However, the information provided by a single transition tuple concerning the probability distributions $P(r_t | s_t, a_t)$ and $P(s_{t+1} | s_t, a_t)$ is too sparse and instead, we propose to learn a causal mapping for each $n^\text{th}$ agent using $P$ samples of the state i.e., $\mathcal{P} = \{(s_t,a_t,r_t,s_{t+1})_p|p=1,\dots,P\}_n$ where data-set $\mathcal{P}$ ought to provide sufficient information to the reward and state transition functions (i.e., the density of $\mathcal{P}$ does not resemble a delta function).
\subsection{Model Parameter Extraction via Independence Enforcement}
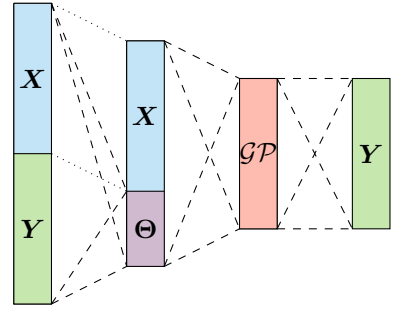
\begin{figure} 
\centering
\begin{tikzpicture}
\draw (0,0) -- (0.5,0) -- (0.5,-4) -- (0,-4) -- (0,0);
\fill [myblue] (0,0) rectangle (0.5,-2);
\node[draw=none,fill=none] at (0.25,-1) {$\bm{X}$};
\fill [mygreen] (0,-2) rectangle (0.5,-4);
\node[draw=none,fill=none] at (0.25,-3) {$\bm{Y}$};

\draw (0+1.5,-0.5) -- (0.5+1.5,-0.5) -- (0.5+1.5,-4+0.5) -- (0+1.5,-4+0.5) -- (0+1.5,0-0.5);
\fill [myblue] (1.5,-0.5) rectangle (2,-2.5);
\node[draw=none,fill=none] at (1.75,-1.5) {$\bm{X}$};
\fill [mypurple] (1.5,-2.5) rectangle (2,-3.5);
\node[draw=none,fill=none] at (1.75,-3) {$\bm{\Theta}$};

\draw (0+1.5+1.5,-0.5-0.5) -- (0.5+1.5+1.5,-0.5-0.5) -- (0.5+1.5+1.5,-4+0.5+0.5) -- (0+1.5+1.5,-4+0.5+0.5) -- (0+1.5+1.5,0-0.5-0.5);
\fill [mypink] (3,-1) rectangle (3.5,-3);
\node[draw=none,fill=none] at (3.25,-2) {$\mathcal{GP}$};

\fill [mygreen] (4.5,-1) rectangle (5,-3);
\draw (4.5,-1) rectangle (5,-3);
\node[draw=none,fill=none] at (4.75,-2) {$\bm{Y}$};

\draw (0,-2) -- (0.5,-2);
\draw (1.5,-2.5) -- (2,-2.5);

\draw (0,0) -- (0.5,0) -- (0.5,-4) -- (0,-4) -- (0,0);
\draw (0+1.5,-0.5) -- (0.5+1.5,-0.5) -- (0.5+1.5,-4+0.5) -- (0+1.5,-4+0.5) -- (0+1.5,0-0.5);
\draw (0+1.5+1.5,-0.5-0.5) -- (0.5+1.5+1.5,-0.5-0.5) -- (0.5+1.5+1.5,-4+0.5+0.5) -- (0+1.5+1.5,-4+0.5+0.5) -- (0+1.5+1.5,0-0.5-0.5);

\draw[dotted] (0.5,0) -- (1.5,-0.5);
\draw[dashed] (0.5,0) -- (1.5,-2.5);
\draw[dashed] (0.5,0) -- (1.5,-3.5);

\draw[dotted] (0.5,-2) -- (1.5,-2.5);

\draw[dashed] (0.5,-4) -- (1.5,-2.5);
\draw[dashed] (0.5,-4) -- (1.5,-3.5);

\draw[dashed] (2,-0.5) -- (3,-1);
\draw[dashed] (2,-0.5) -- (3,-3);
\draw[dashed] (2,-3.5) -- (3,-1);
\draw[dashed] (2,-3.5) -- (3,-3);

\draw[dashed] (3.5,-1) -- (4.5,-1);
\draw[dashed] (3.5,-1) -- (4.5,-3);
\draw[dashed] (3.5,-3) -- (4.5,-1);
\draw[dashed] (3.5,-3) -- (4.5,-3);

\end{tikzpicture}
\caption{An autoencoder interpretation of our model.} \label{autoencoder}
\end{figure}
More concretely, denote $\bm{s}_n = [s_1,\dots,s_P]_n^\top$ as the aggregation of the sampled states for the $n^\text{th}$ agent and similarly, $\bm{r}_n = [r_1,\dots,r_P]_n^\top$ and $\bm{s}_n'=[s_1',\dots,s_P']_n^\top$ as the aggregation of the sampled next states and rewards for the $n^\text{th}$ agent, respectively. We denote the collection of states $\bm{X} \triangleq [\bm{s}_1,\dots,\bm{s}_N]^\top$ as the observed cause and the set of next states $\bm{Y}_s \triangleq [\bm{s}'_1,\dots,\bm{s}'_N]^\top$ or the set of rewards $\bm{Y}_r \triangleq [\bm{r}_1,\dots,\bm{r}_N]^\top$ as the effect depending on if one wishes to extract, respectively, the model parameters $\bm{\xi}$ or $\bm{\omega}$, the matrix collecting the model parameters of each data-point by $\bm{\Theta} = [\bm{\theta}_1,\dots,\bm{\theta}_N]^\top$ and the random variable that contributes to the effect by $\tilde{\bm{X}} = [\bm{X}, \bm{\Theta}]$. Emulating the formulation of ANM-MMs, we apply Gaussian process latent variable model (GP-LVM) \cite{Lawrence:2005:PNP:1046920.1194904} for mapping observation to latent, but more specifically, we use a back-constrained version of GP-LVM \cite{Lawrence:2006:LDP:1143844.1143909} that introduced a multilayer perceptron (MLP) mapping which preserves local distances between observation space and latent space. Denote $\bm{Z} = [\bm{X}, \bm{Y}]$, then let $\bm{\Theta}$ be the output of the constraining MLP $E$ with inputs $\bm{Z}$, parameterized by the weights $\bm{w}$, i.e. $\bm{\Theta} \triangleq E_{\bm{w}}(\bm{Z})$. Following \cite{Hu:2018:CIM:3327345.3327427}, we learn the model parameters by mapping $\bm{\Theta}$ back to observation space using a Gaussian process (GP) and therefore, denoting $\tilde{\bm{X}} = [\bm{X}, \bm{\Theta}]$, the log-likelihood loss of the observations is given by \begin{multline*}
\mathcal{L}(\bm{\Theta}|\bm{X}, \bm{Y}, \beta) = -\frac{DN}{2}\ln(2 \pi) - \frac{D}{2} \ln \big (|\tilde{\bm{K}}| \big ) \\- \frac{1}{2} \text{Tr}\big( \tilde{\bm{K}}^{-1} \bm{Y} \bm{Y}^{-1}  \big),
\end{multline*}where $\tilde{\bm{K}}$ denotes a higher dimensional kernel mapping of the inputs $\tilde{\bm{X}}$ and in this paper, we adopt the radial basis function (RBF) kernel. In summary, our architecture may be interpreted as an autoencoder consisting of a MLP encoder and a GP decoder as illustrated in Figure \ref{autoencoder}.

The model parameters cannot be found by maximizing just the log likelihood alone since ANM-MMs additionally require the independence between $X$ and $\theta$. To this end, we apply independence enforcement via HSIC, whose empirical estimator based on a sampled dataset $\bm{\mathcal{D}} \triangleq \{(\bm{x}_n, \bm{y}_n)\}_{n=1}^N$ is given by $$\text{HSIC}_b(\bm{\mathcal{D}}) = \frac{1}{N^2}\text{Tr}(\bm{KHLH})$$ where $\bm{H}=\bm{I}-\frac{1}{N}   \vec{\bm{1}}\vec{\bm{1}}^\top$, $\vec{\bm{1}}$ is a $N \times 1$ vector of ones and both $\bm{K}$ and $\bm{L}$ are kernel mappings. Incorporating the HSIC term, we arrive at the final loss objective given by \begin{multline} \label{loss}
    \underset{\bm{\Theta}, \Omega}{\argmin} \text { }  \mathcal{J}(\bm{\Theta}) = \underset{\bm{\Theta}, \Omega} {\text{arg min}}[-\mathcal{L}(\bm{\Theta} | \bm{X}, \bm{Y}, \Omega) \\+ \lambda \log \text{HSIC}_b(X, \bm{\Theta})]
\end{multline} where $\lambda$ controls the importance of the HSIC term and $\Omega$ is the set of all hyperparameters including all of the kernel hyperparameters. $\mathcal{J}(\bm{\Theta})$ can be minimized by any stochastic gradient-based optimizer and in this work, we use the scaled conjugate gradient algorithm \cite{scg}.
\subsection{Soft Mechanism Clustering for Similarity-Based Data Sharing}
For each observation pair $(\bm{x}_n, \bm{y}_n)$, there is an associated model parameter $\bm{\theta}_n$ which characterizes its underlying data generating mechanism. Furthermore, common data generating mechanism would have similar $\bm{\theta}$ and hence are identifiable with respect to $\bm{\theta}$. In the windy pendulum example, the model parameters may correspond to various wind strengths experienced by the agents and under our framework, similar wind strengths may be grouped with respect to similar model parameters. To identify common data generating mechanism, we employ a soft clustering mechanism in the form of Gaussian mixture model (GMM) clustering over $\bm{\Theta}$. Given $C$ components of Gaussian mixtures, we apply the EM algorithm to assign each $\bm{\theta}_n$ its probability of being in each of the $C$ centroids.

At convergence of the EM algorithm each model parameter is associated with a vector $v_n \in \mathbb{R}^C$ lying in the probability simplex (i.e., $v_n \succcurlyeq 0, \|v_n\|_1 = 1$) where the $c^\text{th}$ entry of $v_n$ represents the probability that $v_n$ belongs to the $c^\text{th}$ component for all $c \in \{1,\dots,C\}$. Each $v_n$ is treated as a feature vector and together are used to define a pairwise similarity measure given by a RBF kernel, i.e., \begin{equation*}
\begin{aligned}
\mathcal{K} &\triangleq K(v_m, v_n) \\ &= \text{exp}\bigg(-\frac{\|v_m - v_n\|_2^2}{2}\bigg)~\forall m,n \in \{1,\dots,N\}.
\end{aligned}
\end{equation*}The degree in which data is shared for the $n^\text{th}$ agent across all the other agents is then dictated by how similar the $n^\text{th}$ agent's environment is compared to the environments of all other agents. For this, we first calculate $\hat{\mathcal{K}}$ which is the result of row normalizing $\mathcal{K}$ such that each row sums to unity, i.e., $\hat{\mathcal{K}} =  \Big[\frac{\mathcal{K}_1^\top}{\sum_{q=1}^N \mathcal{K}_{1,q}}, \dots,\frac{\mathcal{K}_N^\top}{\sum_{q=1}^N \mathcal{K}_{N,q}}\Big]^\top$. Next, upon training the $n^\text{th}$ agent with batch size $B$, a minibatch is constructed by aggregating the data obtained via random sampling from all other agents' replay buffers where the amount of data sampled from the $q^\text{th}$ agent's replay buffer is given by $\bar{\mathcal{K}}_{n,q}$ where $\bar{\mathcal{K}}_n = \text{round}(B \hat{\mathcal{K}}_n)$ and $\text{round}(\cdot)$ is the element-wise rounding operator.
\subsection{Coordinated Exploration Heuristic}
Under the CRL setting, there are benefits in coordinating the selection of actions of a group of agents. For example, intuition tells us that two similar agents should take different actions if they also happen to be in similar states as this would have the desirable effect of encouraging diverse exploration across the group of agents. In this work, we propose a simple action selection heuristic which achieves just that. The heuristic works as follows: at the beginning of each episode, each agent $n$ samples a $\mu_n \sim \mathcal{N}(0, \sigma^2)$ which acts as the mean to an Ornstein-Uhlenbeck process whose noise is used to perturb the actions of agent $n$. At each time-step $t$, the noise scale $\sigma$ gets anneal in an $\epsilon$-greedy-like fashion so that its average perturbance approaches zero as time approaches infinity. Informed data-sharing then allows similar agents to be cognizant of state spaces which it may otherwise be oblivious to. Despite its simplicity, we demonstrate in the experiments its significance in the sparse reward setting. \begin{algorithm}[t]
\caption{Causal Coordinated Concurrent Reinforcement Learning}
\label{ouralgo}
\begin{algorithmic}[1]
    \STATE Initialize environments $E_n$, DDPGs $D_n$, and replay buffers $\mathcal{B}_n$ for $n=1,\dots,N$
    \STATE Initialize scale factors $\sigma_1$, $\sigma_2$
    \STATE Generate cause $\bm{X}$ by uniformly sampling from $[-s_\text{min}, s_\text{max}]$
    \STATE Generate uniformly random policy $\pi$
    \FOR{each $n^\text{th}$ agent}
        \STATE Generate effect $\bm{Y}_n$ by evaluating $\bm{X}_n$ with $\pi$ \COMMENT{i.e., rewards or next states depending on extracting $\bm{\xi}$ or $\bm{\theta}$}
    \ENDFOR
    \STATE Determine latent model parameters $\bm{\Theta}$ by optimizing Loss (\ref{loss})
    \STATE Determine $\hat{\mathcal{K}}$ by fitting GMM clustering then row-normalizing
    \FOR{\texttt{e} number of \texttt{epochs}}
    \STATE Initialize $\text{OU}(\mu_n, \sigma_2)$, $\mu_n \sim \mathcal{N}(0, \sigma_1^2)$ for $n=1,\dots,N$
    \WHILE{epoch is not over}{}
        \FOR{each $n^\text{th}$ agent}
            \STATE Take action $a + \epsilon$ with $a$ according to $D_n$ and $\epsilon \sim \text{OU}(\mu_n, \sigma_2)$ in $E_n$
            \STATE Cache experience tuple $(s,a,r,s')$ to $\mathcal{B}_n$
        \ENDFOR
        \STATE Anneal $\sigma_1$, $\sigma_2$
        \FOR{each $n^\text{th}$ agent}
            \STATE Create mini-batch $\mathcal{B}_M$ by sampling from all other agents' $\mathcal{B}$ according to $\bar{\mathcal{K}}_n$
            \STATE Update critic and actor network of $D_n$ with $\mathcal{B}_M$
        \ENDFOR
    \ENDWHILE
    \ENDFOR

\end{algorithmic}
\end{algorithm}
\subsection{A Connection to Transportability}
In this section we provide an intuition by drawing connections to the causality model of \cite{10.1093/biomet/70.1.41} and transportability \cite{Bareinboim7345} in causal inference. First we define our notation: Let $Q_t$ be query at time $t$ where query is next state prediction, $b(E)$ the balancing score for a covariate $E$ where in our context, $e \in E$ is a individual realization in the set of MDPs, $\pi$ the policy and $g(E) = \mathbb{P}(\pi | b(E))$ the propensity score.

According to \cite{10.1093/biomet/70.1.41}, a score $b(E)$ can act as a balancing score for covariates $E$ if it satisfies the following conditions:\begin{enumerate}
    \item a balancing score must be ``finer'' than the propensity score and
    \item our ``treatment assignment'', or propensity score, must be strongly ignorable given the balancing score.
\end{enumerate}
\begin{assumption}\label{assume}
All realizations of an environment $e \in E$ vary only by parameter $\theta$ and are otherwise the same.
\end{assumption} This follows from our problem definition and is illustrated as well in our windy pendulum example where environment only varies by wind represented by $\theta$. Therefore, the difference between environments (i.e., input covariates) can be represented entirely by $\theta$. We show $\theta$ satisfies the conditions above (i.e., is a balancing score).

\subsubsection{$\theta$ Is Finer Than the Propensity Score $g(E)$} because there exists an $f$ such that $g(E) = f(\theta)$ (e.g., $f$ is the exogeneity-injecting wind in our context) and therefore, by Theorem 2 of \cite{10.1093/biomet/70.1.41} $b(E) = \theta$ is a balancing score.

\subsubsection{The Propensity Score is Strongly Ignorable} given $E$ and therefore by Theorem 3 of \cite{10.1093/biomet/70.1.41}, the ``treatment assignment'', in other words, the policy determination must also be strongly ignorable given balancing score $\theta$.

Therefore, having established that $\theta$ satisfies the conditions above, $\theta$ is a fully expressive representation of the covariates. In the case of transportability, we observe a set of next-state predictions on environment instances $e^* \in E$ and we seek to answer the query $Q_t = P(s_{t+1} | \pi(s_t | e^*), E = e^*, s_t) P(\theta, E)$. We draw on the derivation from \cite{Bareinboim7345} to show \begin{equation*}\begin{aligned}
Q_t &= \sum_{\theta} P(s_{t+1}| do(\pi(s_t|e^*)), E=e^*, s_t, \theta)\\
&~~~~~~~~~~~P(\theta | E=e^*, do(\pi(s|e^*)))\\
&= \sum_{\theta} P(s_{t+1}| do(\pi(s_t|e^*)), s_t, \theta)\\
&~~~~~~~~~~~P(\theta | E=e^*, do(\pi(s|e^*)))\\
&= \sum_{\theta} P(s_{t+1}| do(\pi(s_t|e^*)), s_t, \theta) P(\theta | E=e^*)\\
&= \sum_{\theta} P(s_{t+1}| do(\pi(s_t|\theta)), s_t, \theta) P(\theta | e^*)
\end{aligned}
\end{equation*} where $\pi(s | e^*)$ is an alternative notation for the propensity score. The second line follows from ``S-admissibility'', the third line shows independence between $\theta$ and the policy/propensity score (i.e., third rule of do-calculus) and the last line illustrates that $\theta$ is a sufficient balancing score for $e^*$. $\pi(s_t|\theta), \theta)$ can be modelled by any function approximate while $P(\theta | e^*)$ in our work, is estimated using ANM-MM. We stress that it is not the case that any latent variable model may extract a good balancing score as the extracted $\theta$ may be dependent on the policy as well. As a result, we therefore select to use the ANM-MM which employs HSIC to enforce independence between the extracted $\theta$ and the embedded policy information.
\begin{figure*}[t]
  \centering
  \subcaptionbox*{(a) $\mathcal{N}(-4, 0.1^2)$}[.31\linewidth][c]{%
    \includegraphics[width=\linewidth]{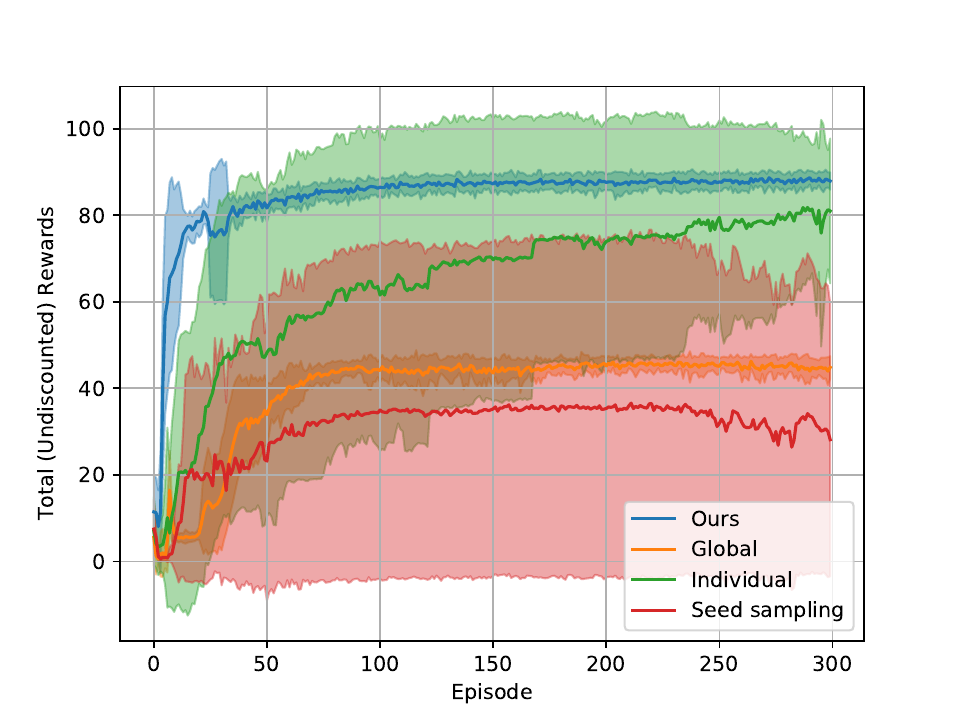}}\quad
  \subcaptionbox*{(b) $\mathcal{N}(-1, 0.1^2)$}[.31\linewidth][c]{%
    \includegraphics[width=\linewidth]{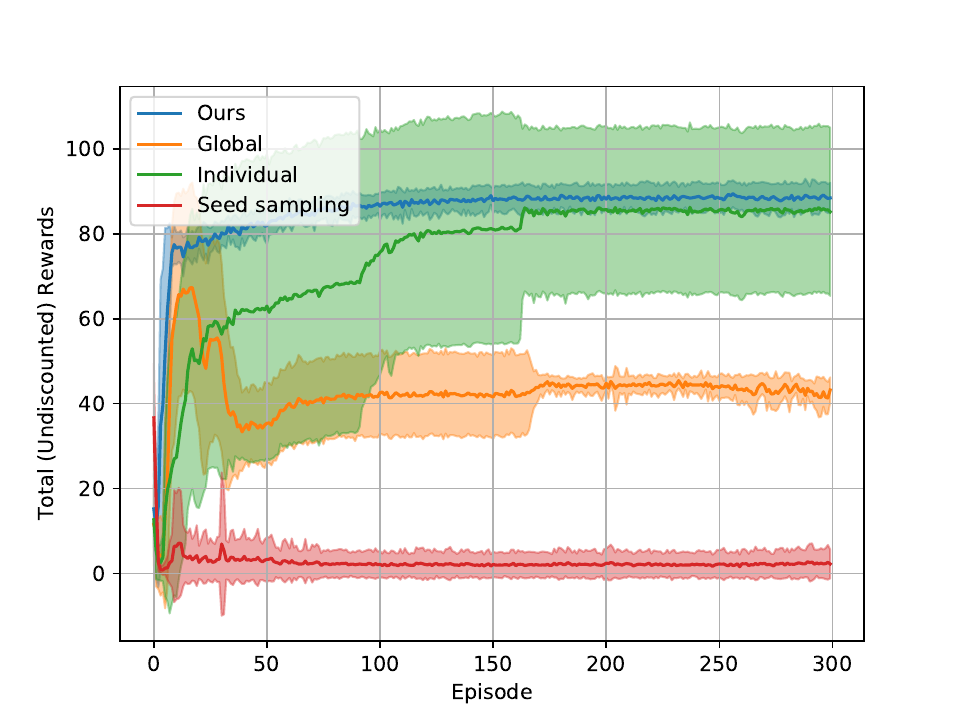}}\quad
  \subcaptionbox*{(c) $\mathcal{N}(4, 0.1^2)$}[.31\linewidth][c]{%
    \includegraphics[width=\linewidth]{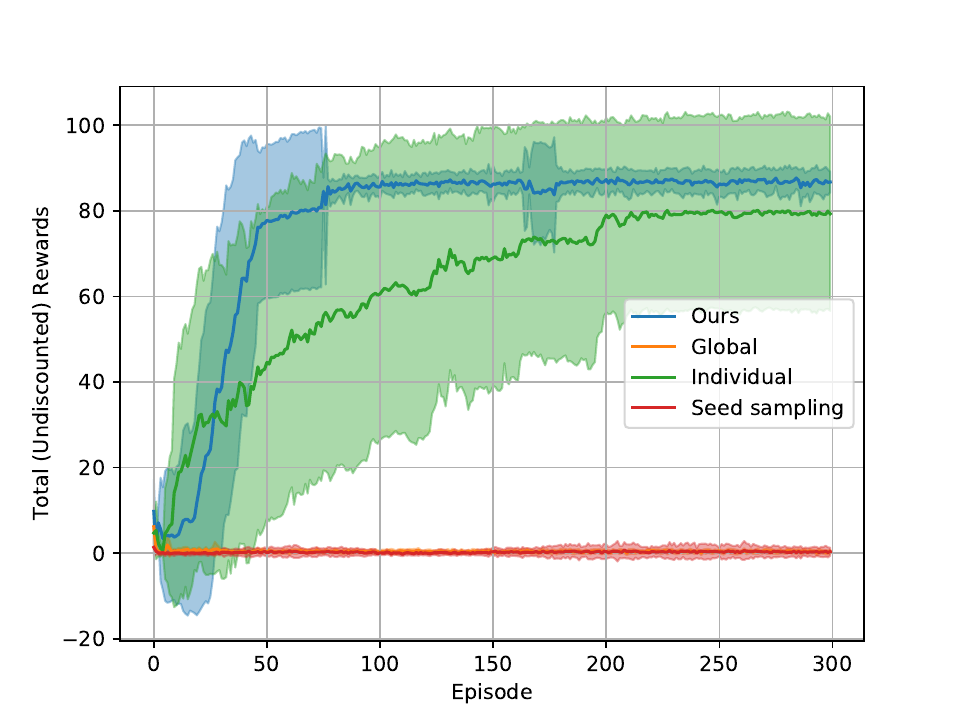}}
  \caption{A comparison between our model vs. three baselines for $s_*$ sampled from a trimodal GMM on the AR task. Shaded region represents one SD of uncertainty from 30 sampled environments.}
  \label{fig1}
\end{figure*}

\begin{figure*}[t]
  \centering
  \subcaptionbox*{(a)}[.23\linewidth][c]{%
    \includegraphics[width=\linewidth]{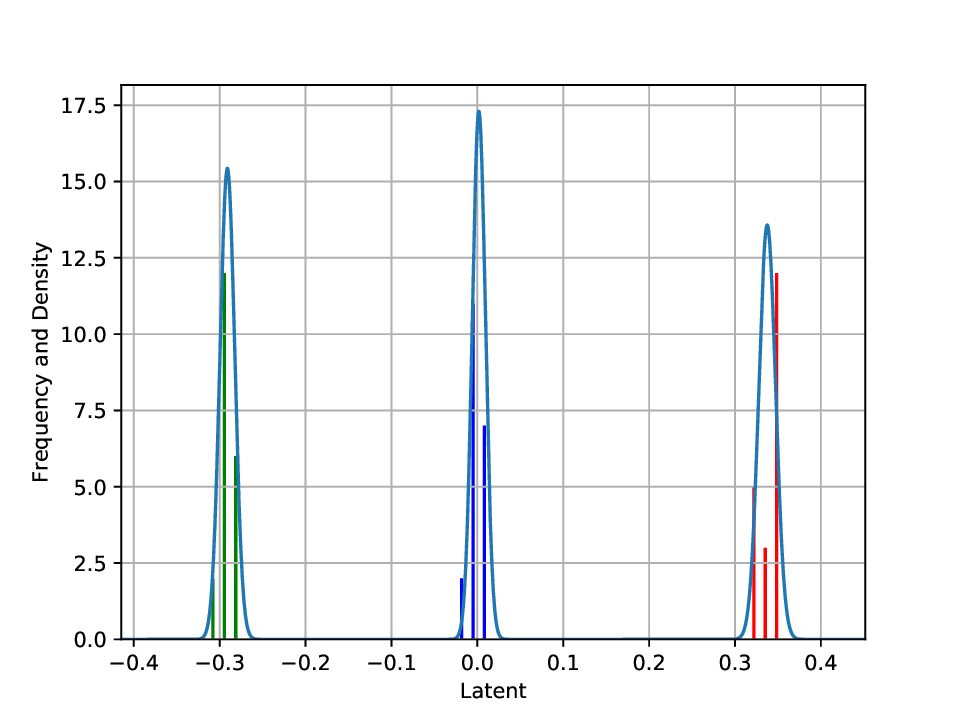}}\quad
  \subcaptionbox*{(b)}[.23\linewidth][c]{%
    \includegraphics[width=\linewidth]{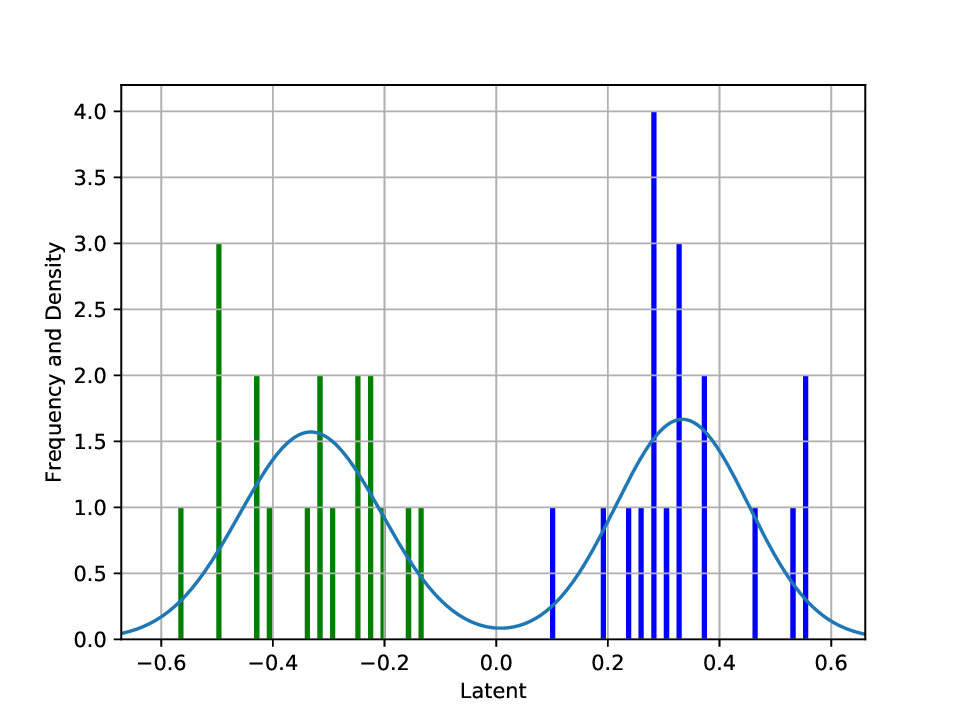}}\quad
  \subcaptionbox*{(c)}[.23\linewidth][c]{%
    \includegraphics[width=\linewidth]{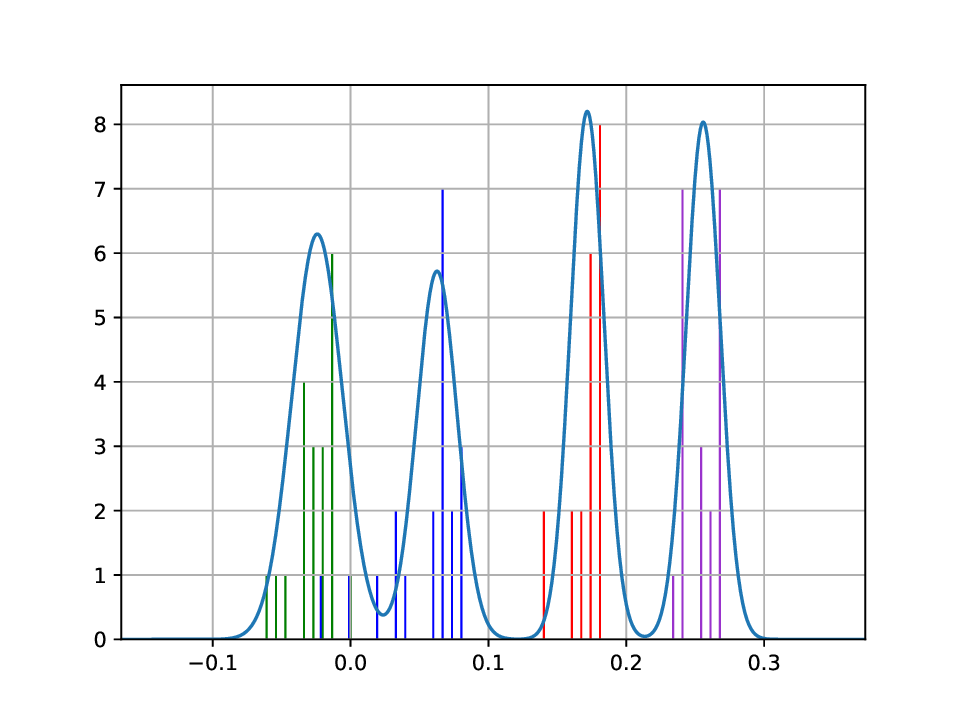}}\quad
  \subcaptionbox*{(d)}[.23\linewidth][c]{%
    \includegraphics[width=\linewidth]{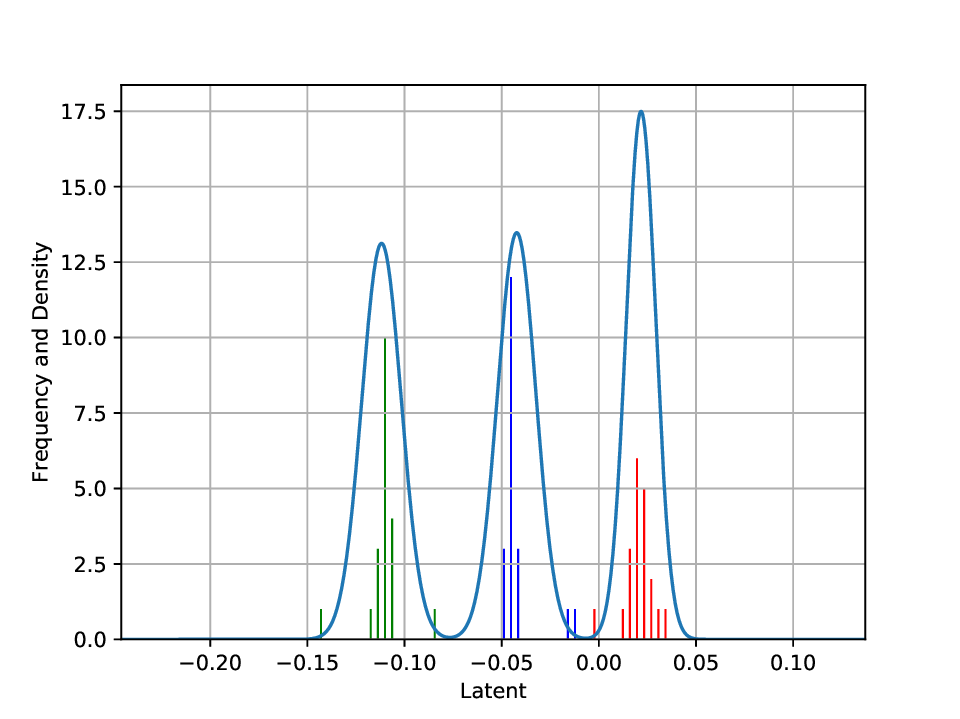}}
    \caption{Histogram of the extracted model parameters fitted with GMM clustering for the (a) AR task, (b) AR task with sparse rewards, (c) windy pendulum task and (d) cart-pole swing-up task. Bars of the histogram are color-coded according to their true data generating mechanism.}
  \label{histograms}
\end{figure*}
\begin{figure*}[t]
  \centering
  \subcaptionbox*{(a) $\mathcal{N}(-20, 0.3^2)$}[.25\linewidth][c]{%
    \includegraphics[width=\linewidth]{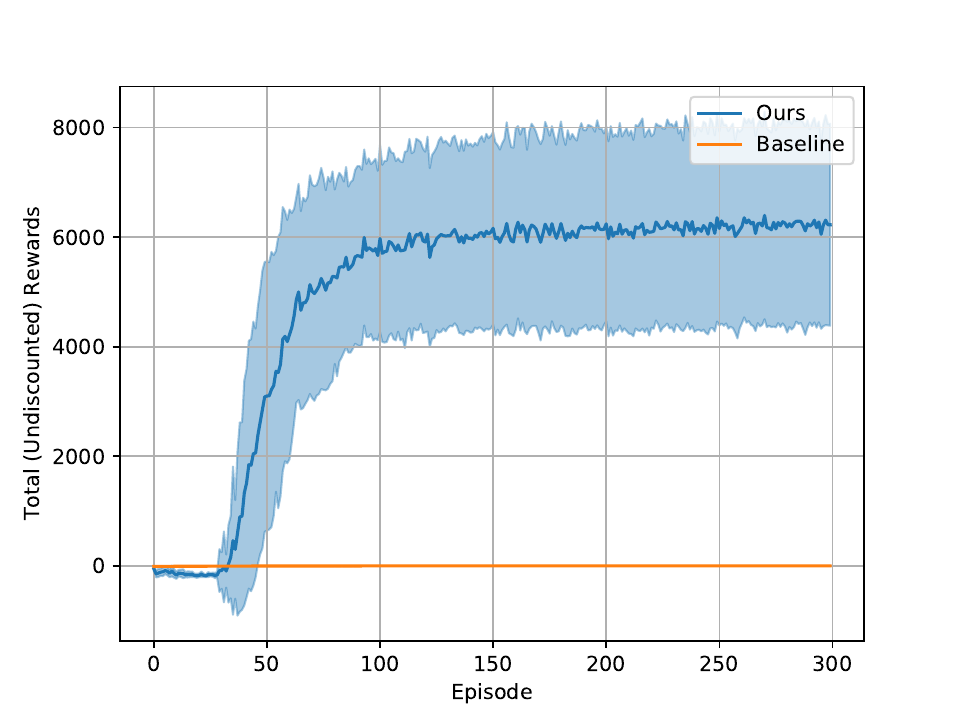}}\quad
  \subcaptionbox*{(c) Baseline at epoch 1}[.25\linewidth][c]{%
    \includegraphics[width=\linewidth]{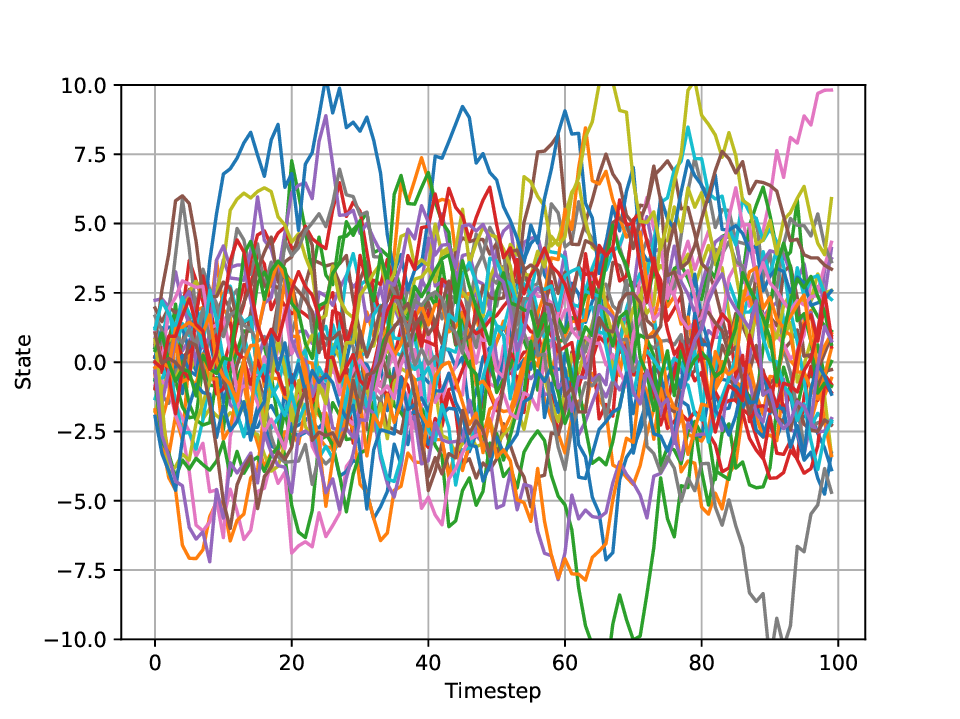}}\quad
  \subcaptionbox*{(e) Ours at epoch 1}[.25\linewidth][c]{%
    \includegraphics[width=\linewidth]{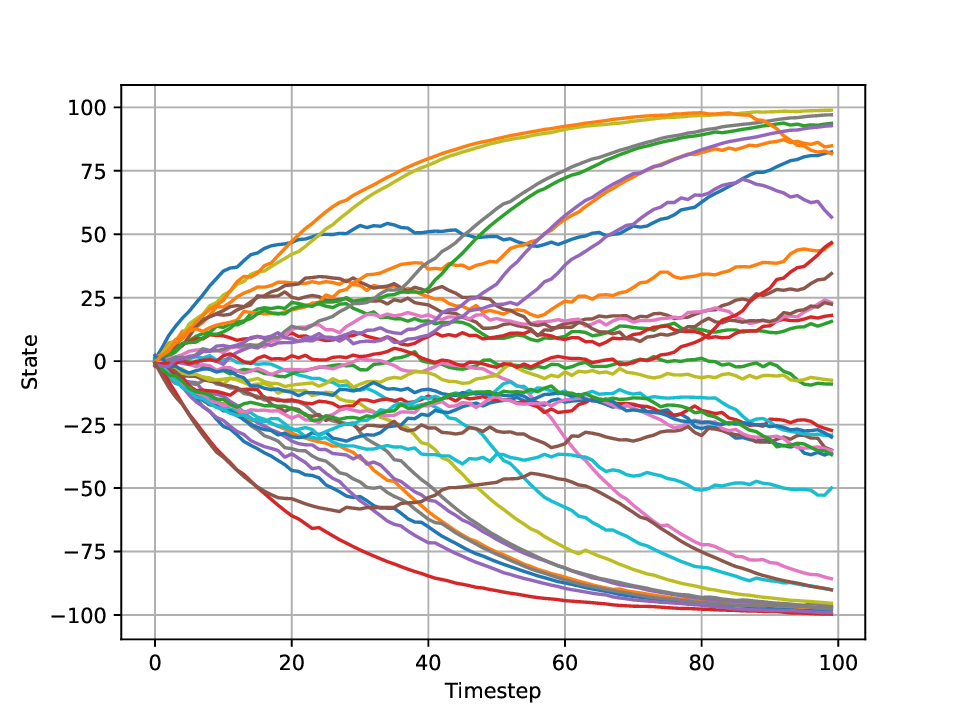}}

  \bigskip

  \subcaptionbox*{(b) $\mathcal{N}(20, 0.3^2)$}[.25\linewidth][c]{%
    \includegraphics[width=\linewidth]{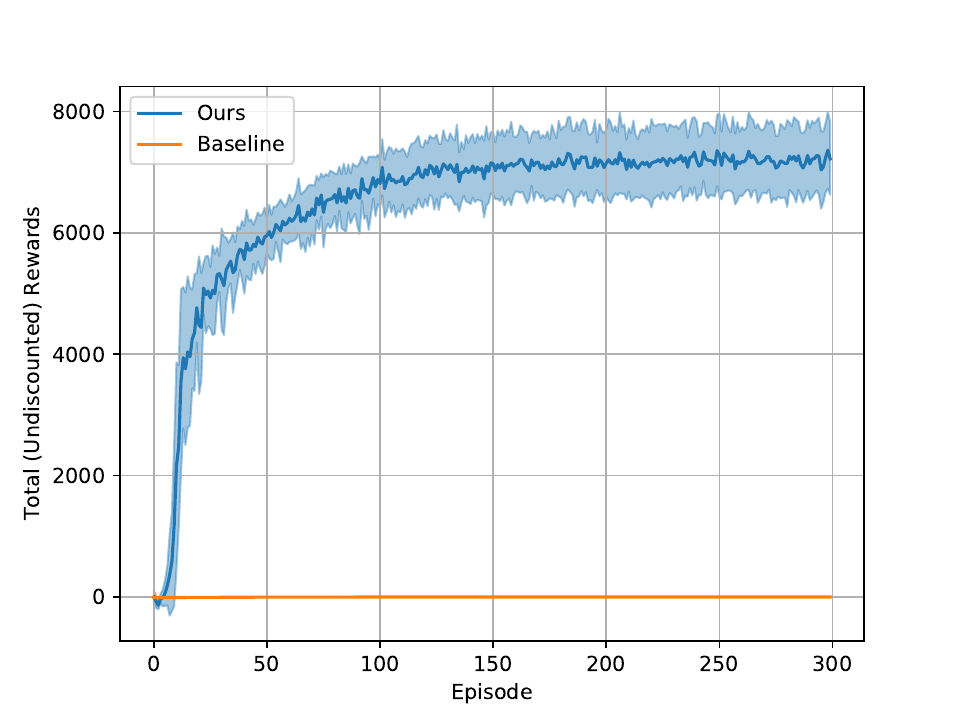}}\quad
  \subcaptionbox*{(d) Baseline at epoch 300}[.25\linewidth][c]{%
    \includegraphics[width=\linewidth]{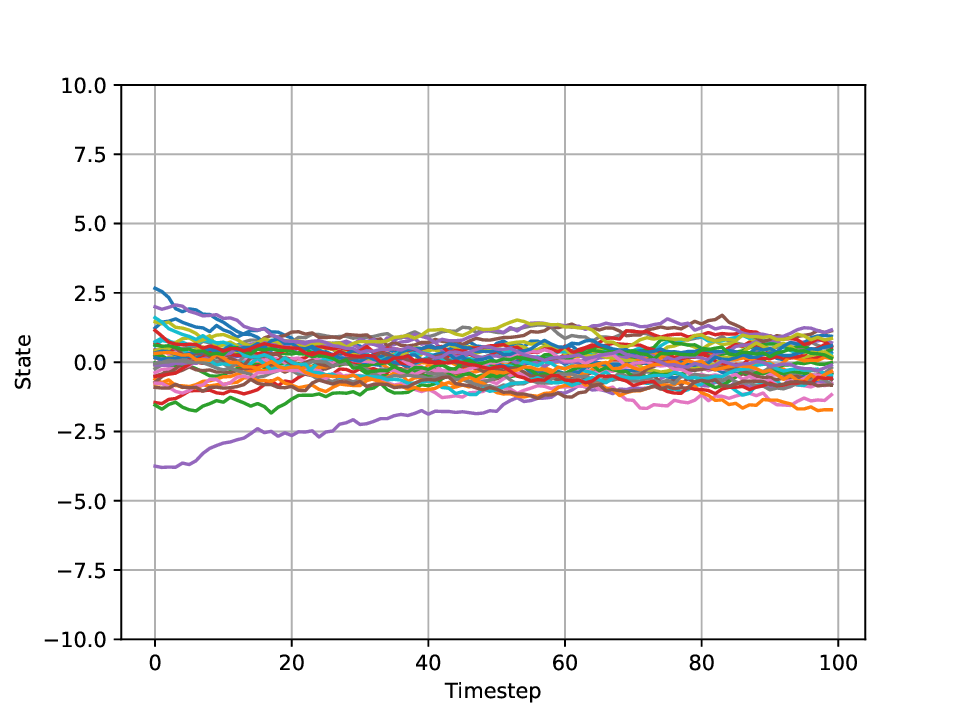}}\quad
  \subcaptionbox*{(f) Ours at epoch 300}[.25\linewidth][c]{%
    \includegraphics[width=\linewidth]{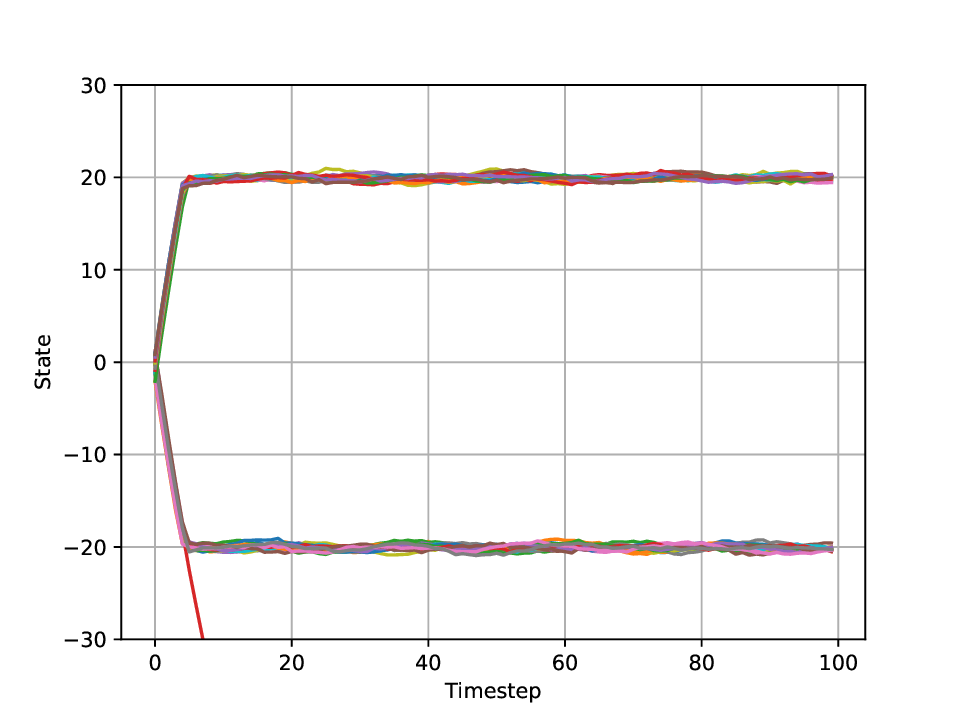}}
  \caption{(a) and (b): A comparison between our coordinated action selection heuritistic vs. a CRL agnostic baseline for $s_*$ sampled from a bimodal GMM on the sparse reward AR task. Shaded region represents one SD of uncertainty from 18 sampled environments. (c)-(f): State trajectories of the AR processes for various instances.}
  \label{fig3}
\end{figure*}
\section{Experiments}
We describe our experimental setups and the results on an autoregressive (AR) task, an AR task with sparse rewards, pendulum and cart-pole swing-up task. In all of our experiments, we assume continuous states and actions and therefore, opt for the actor-critic algorithm, deep deterministic policy gradient (DDPG) \cite{DBLP:journals/corr/LillicrapHPHETS15} for policy learning. In every experiment, each instantiation of DDPG had the following settings: a critic network consisting of two hidden layers of 256 units which receives the states and actions as inputs, another hidden layer of 128 units and a single output unit parameterizing the state-action value. An actor network consisting of a hidden layer of 256 units, another hidden layer of 128 units and finally, an output layer with action dimension number of units. All units use the ReLU activation function except for the output of the critic network which uses linear activation and the output of the actor network which uses $\tanh(\cdot)$ activation to bound the action space.
\subsection{Autoregressive Task}
We conduct some preliminary experiments on AR tasks as a proof of concept of the parameter extraction capabilities and the faster learning speeds of our algorithm. The AR model used in this experiment is defined by the state evolution equation $s_{t+1} = \phi s_t + a_t + \epsilon$ where $\phi$ is a constant set to 0.95 and $\epsilon \sim \mathcal{N}(0, 0.1^2)$ is the noise parameter. The reward is defined as $r_t = \text{exp}(-|s_t - s_*|)$, in other words, the goal of the agent is to drive and maintain its state as close as possible to $s_*$. Initial state $s_0$ is sampled from a standard normal distribution. We examine the model parameter extraction capabilities of our algorithm based on variations in the reward function and to this end, we create three groups of 20 MDPs where the first group is instantiated with a reward function $s_* \sim \mathcal{N}(-4, 0.1^2)$, the second group with $s_* \sim \mathcal{N}(-1, 0.1^2)$ and the last group with $s_* \sim \mathcal{N}(4, 0.1^2)$. The set of states representing the cause, that is, $\bm{X}$ is generated by uniformly sampling 100 state samples in the interval $[-10, 10]$ and each agent shares a common cause that is perturbed slightly with Gaussian noise (so as to prevent singular matrices during the optimization procedure). To generate the corresponding effect $\bm{Y}$, we first generate a data set $\mathcal{D}_n = \{(s,a,r,s')_m|m=1,\dots,M\}$ of transition tuples for each agent $n$ using random policies. Then for each $s \in \bm{X}$, the corresponding cause for agent $n$ is defined by $y_n(s) \triangleq \mathcal{D}_n(m_*)[r]$ where $m_* = \argmin_m \| s - \mathcal{D}_n(m)[s] \|_2^2 $ and $\mathcal{D}_n(m)[x]$ denotes element $x \in \{s, a, r, s'\}$ of the $m^\text{th}$ tuple of $\mathcal{D}_n$. Figure \ref{histograms}(a) depicts a histogram of the resulting extracted model parameters along with a fitted GMM clustering. The histogram bars (i.e., hidden model parameters) are color-coded in green, blue and red where each color corresponds to a different data generating mechanism (i.e., mean of target location), which as illustrated, our algorithm is able to cleanly extract and separate.

Next, we compare the learning curves of our proposed data sharing scheme against a baseline where data is naively shared across all agents, a second baseline where concurrency is disregarded and each agent learns from their own experience and a third baseline where the coordinated exploration seed sampling algorithm is used. A batch size of 192, 192, 64 and 192 is used for our algorithm, the first baseline, the second baseline and the third baseline, respectively. The decision for using a larger batch size stems from the rationale that algorithms utilizing data sharing have a larger amount of data available to them and a larger batch size allows models to leverage this. The resulting training curves are depicted in Figure 2(a)-(c). We see that across the three groupings, our replay buffer sharing strategy resulted in faster policy learning as well superior policies compared to the baselines. The results corroborate the idea that naive data sharing poses as a detriment to policy learning due to the fact that target locations are different but benefits can clearly be reaped with judicious data sharing when contrasted with policies learned with no data sharing. Similar to naive data sharing, seed sampling also performs poorly with the most likely culprit being the baseline's assumption that all agents interact with underlying identical MDPs.
\subsection{Autoregressive Task with Sparse Rewards}
\begin{figure*}[t]
  \centering
  \subcaptionbox*{(a) $\mathcal{N}(-4, 0.1^2)$}[.23\linewidth][c]{%
    \includegraphics[width=\linewidth]{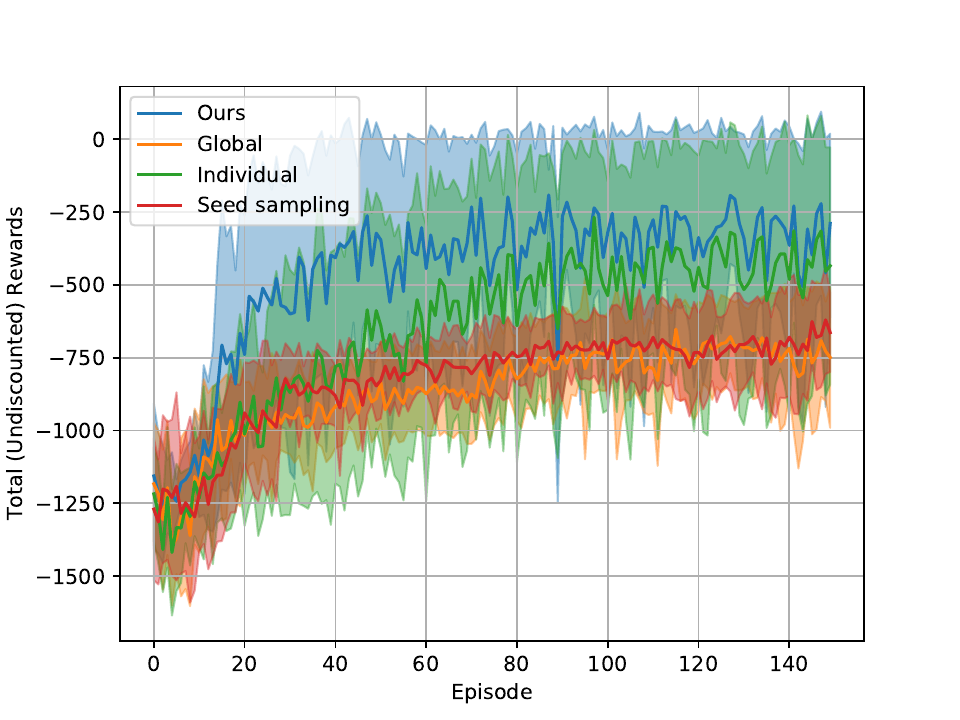}}\quad
  \subcaptionbox*{(b) $\mathcal{N}(-1.5, 0.1^2)$}[.23\linewidth][c]{%
    \includegraphics[width=\linewidth]{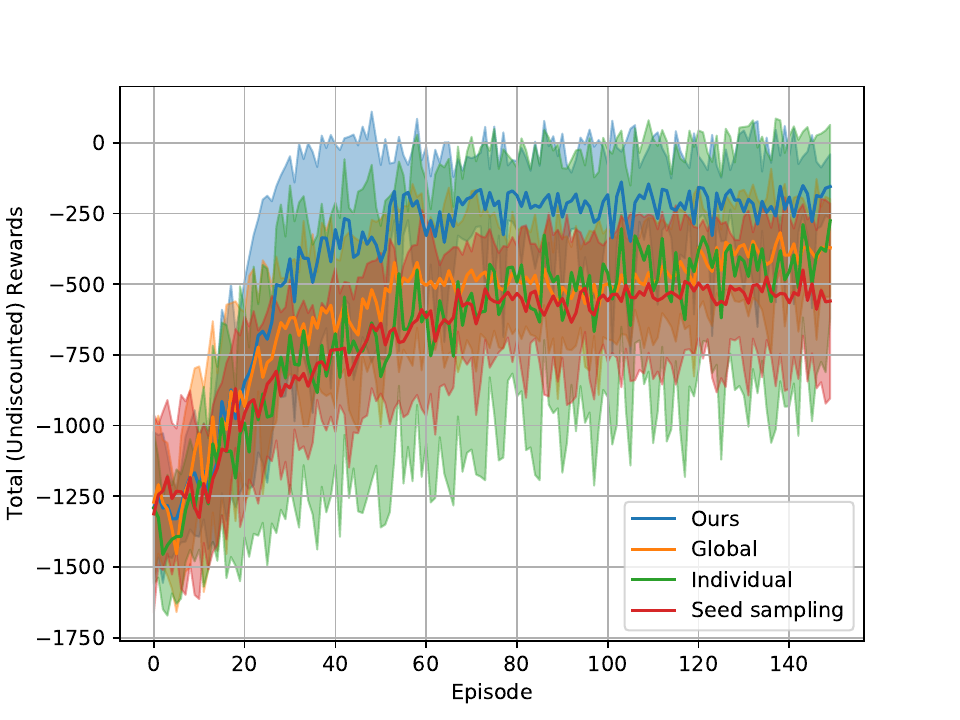}}\quad
  \subcaptionbox*{(c) $\mathcal{N}(1.5, 0.1^2)$}[.23\linewidth][c]{%
    \includegraphics[width=\linewidth]{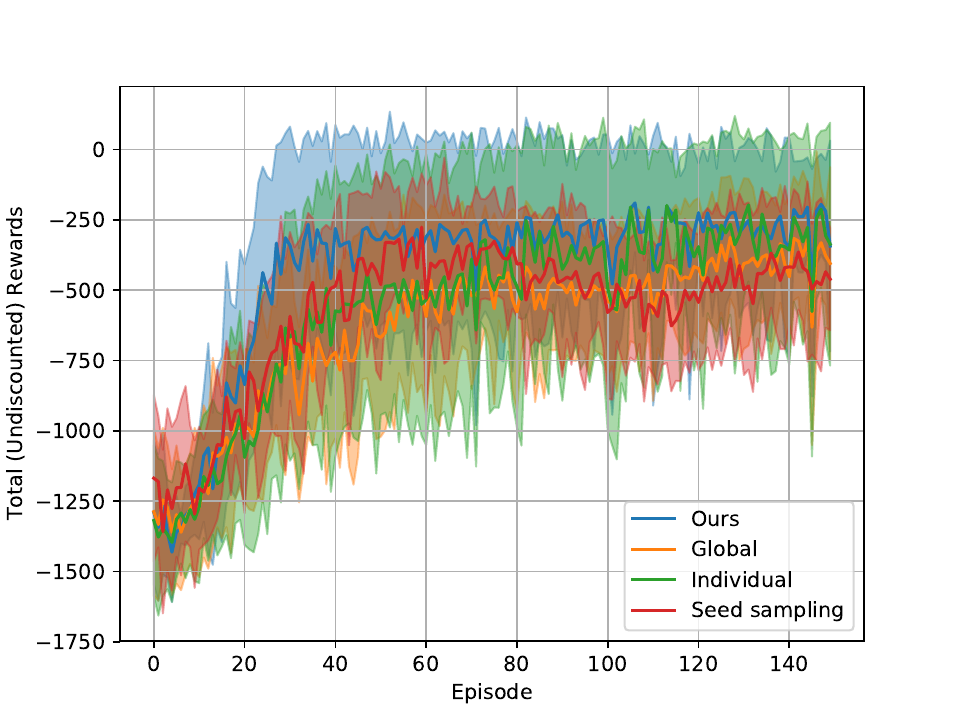}}\quad
  \subcaptionbox*{(d) $\mathcal{N}(4, 0.1^2)$}[.23\linewidth][c]{%
    \includegraphics[width=\linewidth]{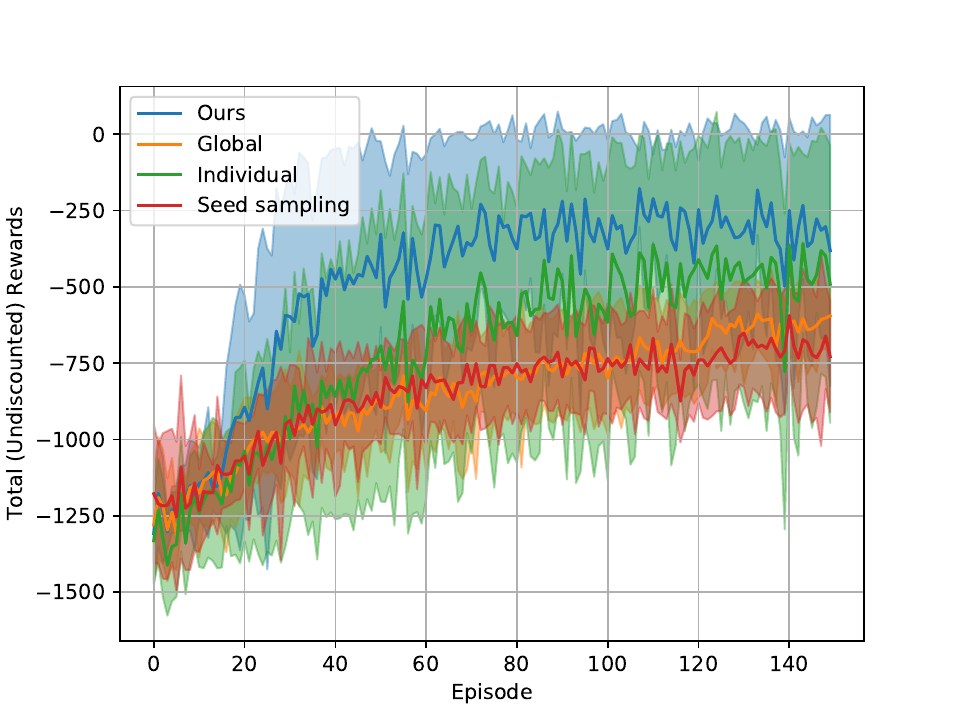}}
    \caption{A comparison between our model vs. three baselines for wind strengths sampled from a multimodal GMM on the windy pendulum task. Shaded region represents one SD of uncertainty from 20 sampled environments.}
  \label{pendulum}
\end{figure*}
\begin{figure*}[t]
  \centering
  \subcaptionbox*{(a) $\mathcal{N}(7.82, 0.1^2)$}[.31\linewidth][c]{%
    \includegraphics[width=\linewidth]{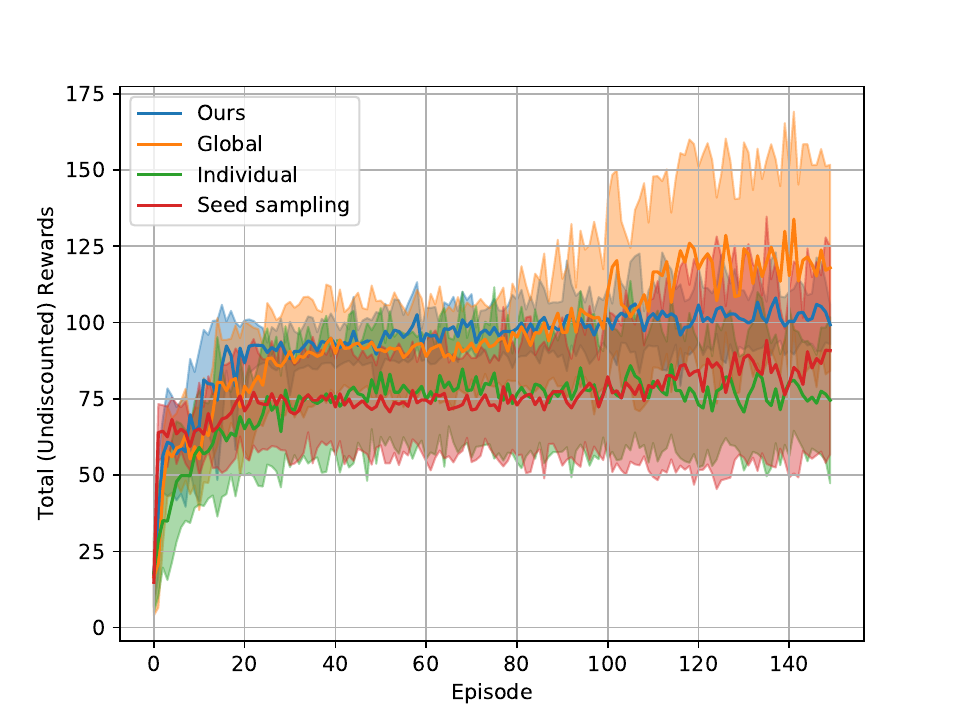}}\quad
  \subcaptionbox*{(b) $\mathcal{N}(11.82, 0.1^2)$}[.31\linewidth][c]{%
    \includegraphics[width=\linewidth]{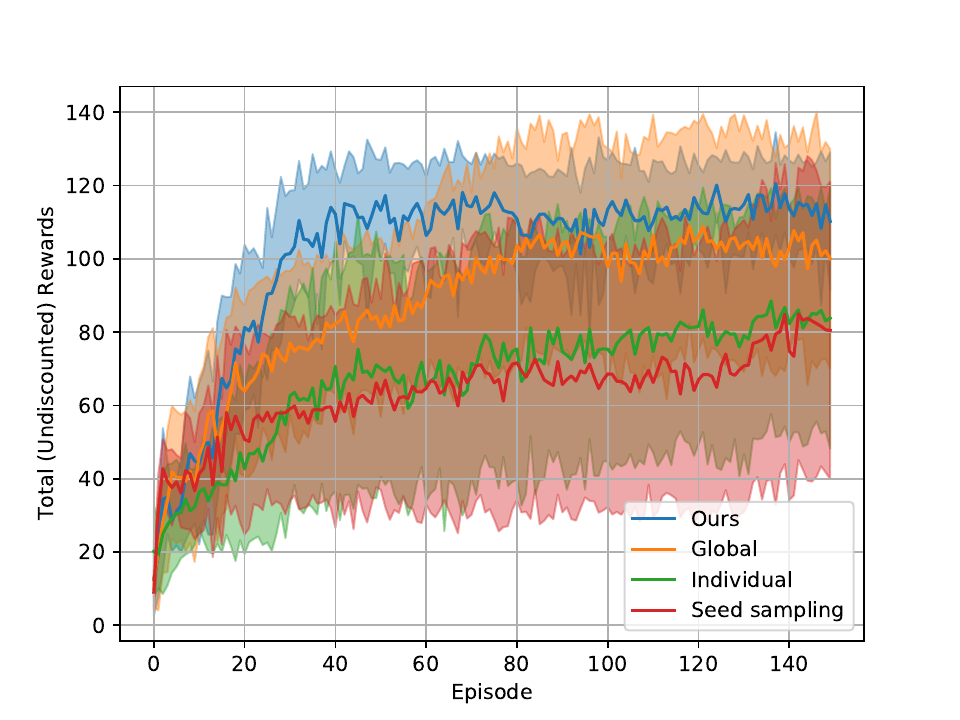}}\quad
  \subcaptionbox*{(c) $\mathcal{N}(15.82, 0.1^2)$}[.31\linewidth][c]{%
    \includegraphics[width=\linewidth]{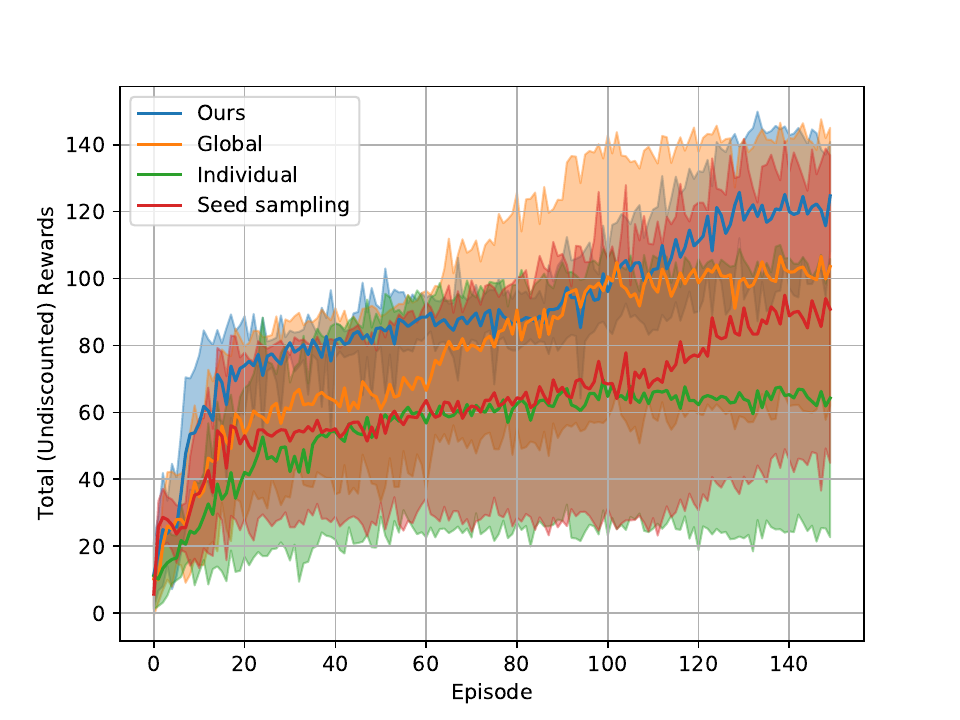}}
    \caption{A comparison between our model vs. three baselines for gravity strengths sampled from a trimodal GMM on the cart-pole swing-up task. Shaded region represents one SD of uncertainty from 20 sampled environments.}
  \label{cartpoleswingup}
\end{figure*}
Under a CRL setting it is intuitive to coordinate the actions of each agent in order to efficiently explore the state space as quickly as possible. The benefits of this are most pronounced under sparse reward settings and we demonstrate the superior performance of our action selection heuristic coupled with our data sharing strategy under this setting. We use the same AR state evolution equation in the previous experiments but in order to make the task exhibit sparse rewards, we modify the reward function so that there is a penalty incurred for making an action proportionate to the magnitude squared of the action and the agent receives a large reward for moving to the target, i.e., $r_t = 100 e^{-|s - s_*|} - \frac{a_t^2}{10}$. We create two groups of 18 MDPs where the first group is instantiated with a reward function where $s_* \sim \mathcal{N}(-20,0.3^2)$ and the second group with $s_* \sim \mathcal{N}(20,0.3^2)$. The set of states representing the cause is generated by uniformly sampling 100 states in the interval $[-25, 25]$ and the corresponding effect is generated using the same procedure as the previous experiment. Figure \ref{histograms}(b) depicts the extracted model parameters and a fitted GMM clustering, showing clear separation between the two clusters.

We compare our algorithm to a baseline algorithm which only shares data based on environment similarity but does not coordinate their exploration efforts (i.e., each agent executes their own exploration strategy). The resulting training curves are depicted in Figure \ref{fig3}(a) and Figure \ref{fig3}(b). We see that for both target locations, our algorithm is able to learn a good policy whereas the baseline algorithm is unable to learn anything at all. To investigate why this is the case, we look into the state trajectories of the AR processes of baseline at epoch 1, baseline at epoch 300, ours at epoch 1 and ours at epoch 300 which are depicted in Figure \ref{fig3}(c), Figure \ref{fig3}(d), Figure \ref{fig3}(e) and Figure \ref{fig3}(f), respectively. Each line represents the state trajectory of a single agent. The figures show that for the baseline algorithm, despite exploring the environment during the initial episodes, each individual's exploration strategy is insufficient in discovering the location of the large reward payoff and instead settled with a sub-optimal policy which involves simply not moving. On the other hand, our action selection heuristic is able to make the agents ``fan out'', allowing them to visit a diverse set of states during the initial episodes, ultimately leading to more successful policies.
\subsection{Pendulum Swing-Up}
In this next experiment we extend our algorithm to the classical pendulum swing-up task subjected to the slight addition of a constant wind strength. We create four groups of 20 MDPs where the first group is instantiated with a wind strength sampled from $\mathcal{N}(-4,0.1^2)$, the second group from $\mathcal{N}(-1.5,0.1^2)$, the third group from $\mathcal{N}(1.5,0.1^2)$ and finally, the last group from $\mathcal{N}(4,0.1^2)$. The state space of the task is given by $\bm{s} \in [\cos(\theta), \sin(\theta), \dot{\theta})]$ where $\theta \in \mathbb{R}$ and $\dot{\theta} \in [-8,8]$ are the angle and angular velocity of the pendulum, respectively. 100 states are uniformly sampled from the state space and their corresponding next state are evaluated using a random policy and the angular velocities of states and next states are taken to be the cause and effect, respectively. Figure \ref{histograms}(c) depicts a histogram of the extracted hidden model parameters along with a fitted GMM clustering and suggests that our algorithm is able to extract and separate in latent space, the various data generating mechanisms corresponding to different wind strengths. Next we examine the performance of our data-sharing strategy against a global sharing baseline, a no sharing baseline and a seed sampling baseline and the resulting learning curves are summarized in Figure \ref{pendulum}.
\subsection{Cart-Pole Swing-Up}
Finally, we test our algorithm on cart-pole swing-up tasks with varying gravity strengths. We create three groups of 20 MDPs where the first group is instantiated with a gravity strength sampled from $\mathcal{N}(7.82, 0.1^2)$, the second group from $\mathcal{N}(11.82, 0.1^2)$ and the last group from $\mathcal{N}(15.82, 0.1^2)$. The state space of the task is given by $\bm{s} \in [x, \dot{x}, \cos(\theta), \sin(\theta), \dot{\theta}]$ where $x$ and $\dot{x}$ are, respectively, the position and velocity of the cart and $\theta$ and $\dot{\theta}$ are, respectively, the angle and angular velocity of the pole. The remainder of the experimental setup is similar to the previous experiment. Figure \ref{histograms}(d) is a histogram of the extracted hidden model parameters along with a fitted GMM clustering and the resulting learning curves are given in Figure \ref{cartpoleswingup}.
\section{Conclusion}
In this paper, we address 1) model identification for non-identical MDPs 2) data-sharing for policy learning and 3) action selection for coordinated exploration, under heterogeneous environments for CRL. We propose an algorithmic framework inspired by the recent work on ANM-MMs where model identification is performed via independence enforcement on latent space. We apply GMM clustering to identify the sources of data generating mechanisms as well as to produce a soft similarity measure that is used to inform a data-sharing strategy. Finally we propound a sampling-based coordinated exploration heuristic that achieves diverse state visitations for a group of concurrent agents. Comparisons and results on an AR task, a sparse AR task and two classical swing-up tasks demonstrate the efficacy of our method. For future work, we plan on extending our causal-based model using variational Bayes methods \cite{DBLP:journals/corr/KingmaW13}.

\bibliography{example_paper}
\bibliographystyle{icml2020}

\end{document}